\documentclass[a4paper]{cas-sc}
\usepackage[numbers]{natbib}
\usepackage{multirow}%
\usepackage{amsmath,amssymb,amsfonts}%
\usepackage{amsthm}%
\usepackage{mathrsfs}%
\usepackage[title]{appendix}%
\usepackage{xcolor}%
\usepackage{textcomp}%
\usepackage{manyfoot}%
\usepackage{booktabs}%
\usepackage{algorithm}%
\usepackage{algorithmicx}%
\usepackage{algpseudocode}%
\usepackage{listings}%
\usepackage{appendix} 
\usepackage{amsmath}
\usepackage{multirow}
\usepackage{balance}
\usepackage{amssymb}
\usepackage{amsmath}
\usepackage{longtable}
\usepackage{subfloat} 
\usepackage[switch]{lineno}
\usepackage{xcolor}
\usepackage{graphicx}
\usepackage{subfig}
\usepackage{bbding}
\usepackage{color}
\usepackage{nicematrix}

\def\tsc#1{\csdef{#1}{\textsc{\lowercase{#1}}\xspace}}
\tsc{WGM}
\tsc{QE}
\tsc{EP}
\tsc{PMS}
\tsc{BEC}
\tsc{DE}

\begin{document}
\let\WriteBookmarks\relax
\def\floatpagepagefraction{1}
\def\textpagefraction{.001}
\shorttitle{BCTR: Bidirectional Conditioning Transformer for Scene Graph Generation}
\shortauthors{Hao et~al.}

\title [mode = title]{BCTR: Bidirectional Conditioning Transformer for Scene Graph Generation}    

\author[inst1]{Peng Hao}
\ead{peng1.hao@samsung.com}

\author[inst2]{Weilong Wang}
\ead{wangwl@swufe.edu.cn}

\author[inst1]{Xiaobing Wang}
\ead{x0106.wang@samsung.com}

\author[inst1]{Yingying Jiang}
\ead{yy.jiang@samsung.com}

\author[inst1]{Hanchao Jia}
\ead{hanchao.jia@samsung.com}

\author[inst3]{Shaowei Cui}
\ead{shaowei.cui@ia.ac.cn}

\author[inst5]{Junhang Wei}
\ead{junhang.wei@ia.ac.cn}

\author[inst4]{Xiaoshuai Hao}
\cormark[1]
\orcidauthor{0009-0007-4209-6695}{Xiaoshuai Hao}
\ead{xshao@baai.ac.cn}

\cortext[inst2]{Corresponding author.}

\affiliation[inst1]{organization={Samsung R\&D Institute China–Beijing},
            addressline={No. 12, Taiyangong Middle Road}, 
            postcode={100028}, 
            state={Beijing},
            country={China}}

\affiliation[inst2]{organization={School of Mathematics, Southwestern University of Finance and Economics},
            addressline={555 Liutai Road ,Wenjiang}, 
            postcode={611130}, 
            state={Chengdu},
            country={China}}

\affiliation[inst3]{organization={State Key Laboratory of Multimodal Artificial Intelligence Systems, Institute of Automation, Chinese Academy of Sciences},
            addressline={95 Zhongguancun East Road}, 
            postcode={100190}, 
            state={Beijing},
            country={China}}

\affiliation[inst5]{organization={Laboratory of Cognition and Decision Intelligence for Complex Systems, Institute of Automation, Chinese Academy of Sciences},
            addressline={95 Zhongguancun East Road}, 
            postcode={100190}, 
            state={Beijing},
            country={China}}

\affiliation[inst4]{organization={Beijing Academy of Artificial Intelligence},
            addressline={150 Chengfu Road, Haidian District}, 
            postcode={100084}, 
            state={Beijing},
            country={China}}

\begin{abstract}
Scene Graph Generation (SGG) remains a challenging task due to its compositional property. Previous approaches improve prediction efficiency through end-to-end learning. However, these methods exhibit limited performance as they assume unidirectional conditioning between entities and predicates, which restricts effective information interaction. To address this limitation, we propose a novel bidirectional conditioning factorization in a semantic-aligned space for SGG, enabling efficient and generalizable interaction between entities and predicates. Specifically, we introduce an end-to-end scene graph generation model, the Bidirectional Conditioning Transformer (BCTR), to implement this factorization. BCTR consists of two key modules. First, the Bidirectional Conditioning Generator (BCG) performs multi-stage interactive feature augmentation between entities and predicates, enabling mutual enhancement between these predictions. Second, Random Feature Alignment (RFA) is present to regularize feature space by distilling multi-modal knowledge from pre-trained models. Within this regularized feature space, BCG is feasible to capture interaction patterns across diverse relationships during training, and the learned interaction patterns can generalize to unseen but semantically related relationships during inference. Extensive experiments on Visual Genome and Open Image V6 show that BCTR achieves state-of-the-art performance on both benchmarks.

\end{abstract}





\maketitle


\section{Introduction}

\begin{figure}[!t]
  \centering
  \includegraphics[width=0.9\linewidth]{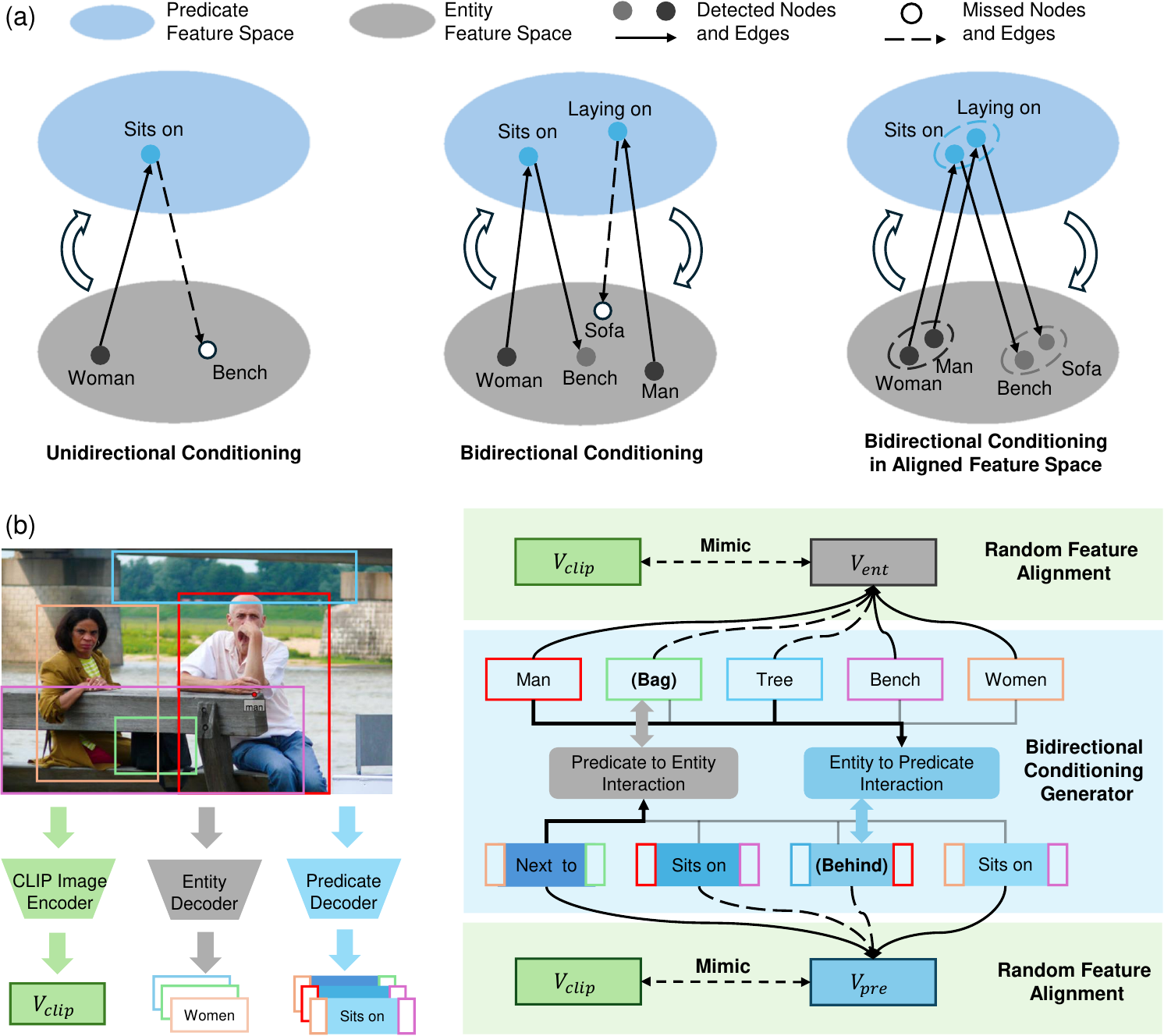}
  \caption{
  \textbf{The motivation and pipeline of BCTR.} (a) Comparison of conditioning approaches in SGG. While bidirectional conditioning generates relationships through mutual feature augmentation even when an entity is missing, it struggles to generalize to unseen categories. Our approach learns bidirectional conditioning in a semantic-aligned space, achieving better generalization to unseen but semantically related relationships. (b) The left panel illustrates the extraction of CLIP, entity, and predicate features from the input image, while the right panel shows how BCTR enhances entity (\textit{e.g.}, "Bag") and predicate (\textit{e.g.}, "Behind") detection through bidirectional interaction in the semantic-aligned space.}
  \label{fig:1}
\end{figure}

\label{sec:intro}

Scene Graph Generation (SGG) aims to enable computers to understand and interpret images by detecting objects and identifying the relationships between them. This process generates structured relationships in the form of triplets (object-predicate-subject). SGG has significant potential for a wide range of applications, including visual question answering~\cite{shi2019explainable}, image captioning~\cite{yang2019auto}, and text-to-image retrieval~\cite{johnson2015image}.

Previous SGG studies can be categorized into two-stage and one-stage approaches. Two-stage methods separate SGG into entity and predicate detection stages, generating $\mathcal{O}(N^2)$ relationship candidates, which demands significant computational resources. Inspired by one-stage object detection~\cite{carion2020end, sun2021sparse}, one-stage SGG methods detect relationships directly from image features~\cite{liu2021fully, teng2022structured}, improving efficiency by avoiding the consideration of all possible pairs. However, these methods lack explicit entity modeling and face challenges with complex relationships within images. To address this, recent one-stage methods~\cite{li2022sgtr, shit2022relationformer} condition predicate prediction on entity features to enhance task performance, while their fixed unidirectional dependence often yields suboptimal results, as entity detection does not benefit from predicate information.

To address this limitation, we propose a bidirectional conditioning factorization in a semantic-aligned space for SGG, enabling efficient and generalizable interactions between entities and predicates, as illustrated in Fig.~\ref{fig:1} (a). Unlike unidirectional conditioning, bidirectional conditioning can predict correct relationships by leveraging feature interactions to infer missing entity/predicate nodes (\textit{e.g.}, Bench). However, due to the long-tail distribution of the SGG dataset, the learned interactions in an unregularized feature space struggle to generalize to unseen but semantically related relationships (\textit{e.g.}, Man-Laying on-Sofa). The core insight of this paper is that learning bidirectional interaction patterns within a semantic-aligned feature space significantly enhances SGG model performance. The benefits lie in two folds. First, bidirectional interaction learning can be facilitated in the semantic feature space during training. Second, the learned interaction patterns can effectively generalize to unseen relationships during inference.

Motivated by this insight, we propose a one-stage SGG model to implement our factorization, dubbed Bidirectional Conditioning TRansformer (BCTR), as illustrated in Fig.~\ref{fig:1} (b). BCTR consists of two core modules: the Bidirectional Conditioning Generator (BCG) and Random Feature Alignment (RFA). The BCG module establishes mutual dependencies between the entity and predicate decoders through two feature interaction mechanisms. The inner interaction utilizes Bidirectional Attention (BiAtt) to enhance information exchange between entities and predicates, while the outer interaction applies iterative refinement to condition current detections on previous estimates, thus improving interactions between predictions. The RFA module is designed to support BCG in learning interaction patterns within a semantic-aligned feature space. By feature distillation with Vision-Language Pre-trained Models (VLPMs), the learned interaction patterns are generalized to unseen but semantically related relationships during inference. We validate BCTR on two SGG datasets: Visual Genome and Open Image V6. Results show that BCTR achieves superior performance compared to existing methods. Our contributions are summarized as follows:

\begin{itemize}
\item We propose a novel bidirectional conditioning factorization within a semantic-aligned space for SGG, enhancing information exchange between predicates and entities by introducing mutual dependence.
\item We present an end-to-end SGG model BCTR to implement our factorization. Specifically, BCG is designed to augment the feature spaces through bidirectional attention mechanisms.
\item We introduce RFA to regularize the feature spaces with VLPMs while preserving diversity for specific tasks. RFA facilitates BCG in learning interaction patterns from long-tail datasets and boosts model performance on tail categories.
\item Extensive experiments on the Visual Genome and Open Image V6 datasets demonstrate that BCTR achieves state-of-the-art performance compared to baselines.
\end{itemize}

The rest of this paper is organized as follows.
We briefly review related works in Section 2. 
In Section 3, we formulate the problem.
We introduce our proposed method in Section 4. 
We then present a variety of experimental results and analyses in Section 5.
Finally, Section 6 concludes this paper.

\section{Related Work}

\subsection{Conditional Dependencies in One-stage SGG}
The previous SGG studies can be divided into two categories: two-stage SGG and one-stage SGG. Two-stage SGG methods~\cite{li2021bipartite} perform entity detection first and then predict the predicate between each pair. Although these methods can capture each possible relationship in the scenario, the predictions are time-consuming and suffer from trivial outputs. In contrast, one-stage SGG methods~\cite{liu2021fully, teng2022structured} directly detect predicates or relationships from image features, avoiding the combination of entities to improve detection efficiency. Inspired by one-stage detection methods~\cite{sun2021sparse, carion2020end}, previous SGG studies have designed Relation Affinity Fields~\cite{liu2021fully} or used query-based detection~\cite{teng2022structured} to directly predict triplets from images. However, these methods perform poorly as they do not utilize entity detection information~\cite{li2022sgtr}. 

Recent works~\cite{shit2022relationformer, cong2023reltr, he2023toward,hao2023mixgen} have explicitly incorporated entity features into relationship detection to improve relationship detection. Nonetheless, these methods either condition predicates on entities~\cite{li2022sgtr, shit2022relationformer} or condition entities on predicates~\cite{desai2022single}, resulting in unidirectional dependencies that limit feature interaction and thus restrict performance gains. Khandelwal et al.~\cite{khandelwal2022iterative} argue that SGG can benefit from dynamic conditioning on the image. By implicitly through a novel joint loss and explicitly through cross-decoder and layer-wise attention mechanisms, their predictions are effectively refined from previous estimates. However, their predicate remains unidirectionally conditioned on the entity, and performance improvement relies on the loss function rather than network architecture. In contrast, our proposed BCG introduces internal bidirectional dependencies, iteratively enhancing features in each round. Additionally, the RFA module enables BCG to learn feature interactions within a semantic-aligned space, improving the generalization of learned interaction patterns to semantically related relationships.

\subsection{Message Passing in SGG}
Message passing aims to boost SGG performance through interactions between predictions and their context. Zhu et al.~\cite{zhu2022scene} categorize it into local (within a triplet) and global (across all elements) message passing. They observe that prediction structures influence the types of message passing: triplet set-based methods~\cite{li2017vip} utilize local message passing, while chain~\cite{xu2017scene}, tree~\cite{tang2019learning}, and fully-connected graph~\cite{li2021bipartite} structures enable global message passing. Global message passing mitigates local ambiguities by integrating contextual information. In contrast to previous methods, we design the BiAtt-based BCG to achieve global message passing within a set structure, leveraging implicit global connections for more efficient information exchange. Moreover, previous message passing generally deploys on two-stage methods~\cite{li2021bipartite}~\cite{xu2017scene}, while the graph generation doesn't influence the entity detections~\cite{khandelwal2022iterative}. Thus, their feature interactions can't be used to optimize entity feature extraction during training. Different from them, BCTR is a one-stage SGG method. The end-to-end mechanism enables the designed feature interaction (BCG) to facilitate the learning of entity detection.

\subsection{External Knowledge in SGG} 
Enhancing SGG performance with external knowledge is a crucial research topic~\cite{zhu2022scene}. Previous SGG works~\cite{yu2017visual, kim2024scene} usually extract statistical information from external textual knowledge (\textit{e.g.}, Wikipedia, ConceptNet). However, statistical information such as co-occurrence frequency is limited in revealing the complex, structured patterns of commonsense, which may lead to poor learning improvement~\cite{lin2022atom}. Generally, using external knowledge to guide feature refinement is another effective approach. Gu et al.~\cite{gu2019scene} proposed a knowledge-based module that improves the feature refinement procedure by reasoning over a collection of commonsense knowledge retrieved from ConceptNet. Zhan et al.~\cite{zhan2019exploring} introduced a novel multi-modal feature-based undetermined relationship learning network. Unfortunately, Zareian et al.~\cite{zareian2020learning} point out that incomplete and inaccurate external commonsense tends to limit task performance. Unlike previous works, BCTR uses VLPMs as a source of external knowledge. By embedding this knowledge through feature distillation, our model learns richer feature representations for scene graph generation, demonstrating superior performance on complex and unseen relationships.

\subsection{Language Prior for Long-Tail Problem} 

The long-tail distribution presents a significant challenge in visual relationship recognition, leading to biased predictions in SGG methods~\cite{zhu2022scene}. The SGG community has developed numerous unbiased SGG methods, with one widely used approach being re-balancing. Re-balancing aims to mitigate the negative impact of the long-tail distribution through re-weighting or re-sampling~\cite{desai2021learning, li2021bipartite}. However, these methods are limited to open-set tasks due to their reliance on statistical priors of datasets. Chang et al.~\cite{chang2021comprehensive} argue that the long-tail problem can be addressed with language priors. Lu et al. \cite{lu2016visual} trained two models, one of which includes a language prior module that projects semantic-like relationships into a more compact embedding space, enabling the model to infer similar visual relationships (\textit{e.g.}, "person riding an elephant" from "person riding a horse"). Their work demonstrates the potential of solving the long-tail problem by aligning the image and language spaces. It is worth noting that language priors are scalable to open-world scenarios compared to statistical priors. However, the enhancement of using language priors to address the long-tail problem is affected by the poor performance of multi-modal feature alignment.

Recently, VLPMs have made breakthroughs in multi-modal tasks~\cite{gan2022vision}, achieving superior performance in aligning image and language features. By distilling features from VLPMs, downstream classifiers can recognize more objects and improve detection results on tail categories~\cite{liao2022gen, he2023toward}. However, research on using language priors in one-stage SGG is still lacking, and current methods fail to address the long-tail problem explicitly~\cite{kundu2023ggt}. Different from previous works, our proposed RFA aligns the visual feature space of SGG with the pre-trained language space to solve the long-tail problem. To our knowledge, RFA is the first method to use a multi-modal model to address data imbalance in SGG, suggesting potential for open-set relation detection.

\section{Problem Formulation}
SGG aims to detect objects and predicates in the input image and represent them as a scene graph $\mathcal{G} = \{\mathcal{V}, \mathcal{E}\}$, where $\mathcal{V}$ and $\mathcal{E}$ denote the sets of vertices and edges, respectively. $\mathcal{V}$ represents all detected objects in the image, while $\mathcal{E}$ comprises the predicates between object pairs. The categories of objects and predicates are defined by the dataset.

Previous one-stage SGG methods typically assume a unidirectional information flow, such as $\mathbf{I}\rightarrow \mathbf{s},\mathbf{o} \rightarrow \mathbf{p}$~\cite{cong2023reltr}, or $\mathbf{I}\rightarrow \mathbf{p}\rightarrow \mathbf{s},\mathbf{o}$~\cite{desai2022single}, limiting mutual benefits between the two predictions. Here, $\mathbf{I}$, $\mathbf{o}$, $\mathbf{s}$, and $\mathbf{p}$ are the abbreviations for image, object, subject and predicate, respectively. Teng et al.~\cite{teng2022structured} formulate SGG as $\mathbf{I}\rightarrow \mathbf{p},\mathbf{s},\mathbf{o}$ to facilitate feature interaction. However, optimizing over the massive compositional triplet space is challenging. This paper proposes a novel factorization for SGG, as shown in Eq.~\ref{eq:3-1}. 
\begin{equation}
    \Pr(\mathbf{E}, \mathbf{P} | \mathbf{I}) = \Pr(\hat{\mathbf{E}}, \widetilde{\mathbf{P}} | \mathbf{I}) \cdot \Pr(\mathbf{E}, \mathbf{P} | \hat{\mathbf{E}}, \widetilde{\mathbf{P}}).
    \label{eq:3-1}
\end{equation}

\noindent where $\mathbf{E}$, $\mathbf{P}$ represent the entity and predicate estimates, respectively. \(\widetilde{}\) and \(\hat{}\) denote the intermediate estimates. The first term reflects $\mathbf{I}\rightarrow \mathbf{s},\mathbf{o}$ and $\mathbf{I}\rightarrow \mathbf{p}$, avoiding the optimization issues caused by the large compositional space. The second term enforces bidirectional dependencies $\mathbf{p}\leftrightarrow \mathbf{s},\mathbf{o}$, allowing the two predictions to benefit from each other. 

After acquiring the predictions of entities and predicates, we formulate SGG as a bipartite graph construction task based on previous work~\cite{li2022sgtr}. Specifically, the predictions of entities and predicates from the image form two node sets $\mathcal{V}_e$ and $\mathcal{V}_p$, respectively. Two directional edge sets $\mathcal{E}_{ep}$ and $\mathcal{E}_{pe}$ are used to connect these node sets, representing the connections from entities to predicates and vice versa. The bipartite graph is then represented as $\mathcal{G}_b=\{\mathcal{V}_e, \mathcal{V}_p, \mathcal{E}_{ep}, \mathcal{E}_{pe}\}$, from which the scene graph of the image can be extracted.

\section{Method}

\begin{figure*}[tb]
  \centering
  \includegraphics[width=1.0\linewidth]{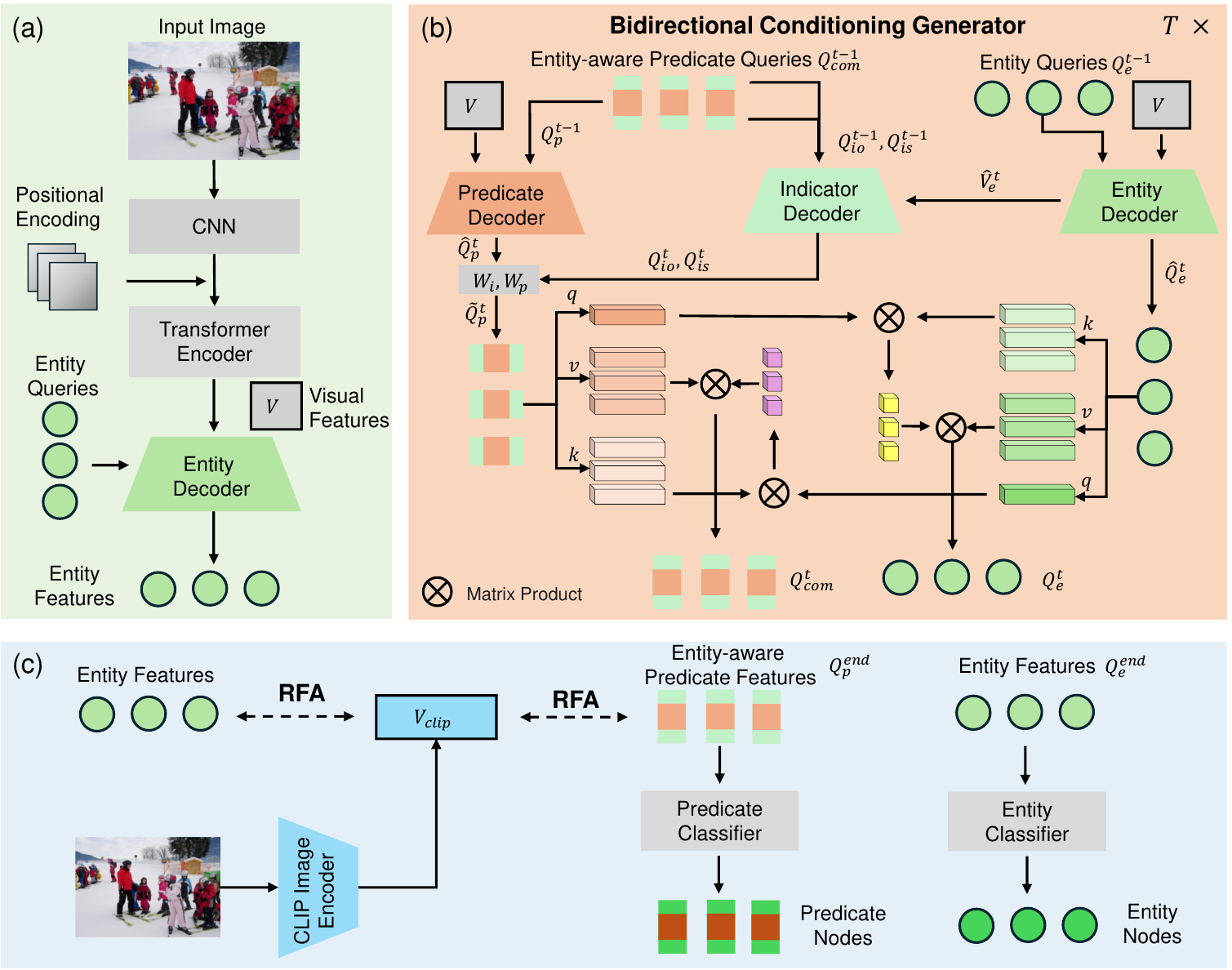}
  \caption{
  \textbf{Overview of the BCTR.}
  (a) Visual and entity features are extracted from the input image. (b) The entity-aware predicate queries and entity queries are iteratively updated through the proposed BCG. (c) During training, the output features from various decoders are regularized by the RFA. Final predictions are generated from these distilled features.}
  \label{fig:overview}
\vspace{-0.5em}
\end{figure*}

This section presents BCTR for implementing the factorization in Eq. \ref{eq:3-1}. The overview of BCTR is shown in Fig. \ref{fig:overview}, which comprises two main modules: BCG and RFA. We detail these components in the sequel. First, we introduce the backbone for feature extraction. Next, we detail BCG, which provides mutual dependence for entity and predicate through BiAtt. RFA is then introduced to regularize the feature spaces with VLPMs and facilitate BCG learning interaction patterns from long-tail datasets. Then, we describe graph assembling, aiming to construct the scene graph from entity and predicate results. Finally, model training and inference are detailed.

\subsection{Feature Extraction}
Inspired by the previous one-stage detection method DETR~\cite{carion2020end}, we utilize a CNN and Transformer to extract features \(\mathbf{V} \in \mathbb{R}^{w \times h \times c}\) from the input image \(\mathbf{I}\), where \(w\), \(h\), and \(c\) represent the width, height, and feature channels, respectively. Since previous one-stage SGG methods have shown that models struggle to capture predicates directly from image features without using intermediate information (\textit{e.g.}, entity features), we further extract entity features \(\mathbf{V}_e \in \mathbb{R}^{N_e \times c}\) from image feature \(\mathbf{V}\) as auxiliary features, as shown in the following:
\begin{equation}
    \mathbf{V}_e = \mathcal{F}_e(\mathbf{V}, \mathbf{Q}_e),
    \label{eq:1}
\end{equation}
\noindent where \(\mathcal{F}_e, \mathbf{Q}_e \in \mathbb{R}^{N_e \times c}\) represent the transformer-based decoder and the learnable queries, respectively, and \(N_e\) denotes the number of queries.

\subsection{Bidirectional Conditioning Generator}

This subsection details the Bidirectional Conditioning Generator, as shown in Fig. \ref{fig:overview}. BCG comprises two interactive branches that take visual features \(\mathbf{V}\) as input and output augmented entity features and entity-aware predicate features, respectively. To improve performance, we introduce the iterative improvement mechanism into BCG. Specifically, we factorize the conditional distribution of entity and predicate according to Eq. \ref{eq:3-1}, which are as follows:
\begin{equation}   
    \Pr(\mathbf{E}^t | \mathbf{I}, \mathbf{E}^{t-1}, \widetilde{\mathbf{P}}^t) = \Pr(\hat{\mathbf{E}}^t | \mathbf{I}, \mathbf{E}^{t-1}) \cdot \Pr(\mathbf{E}^t | \hat{\mathbf{E}}^t, \widetilde{\mathbf{P}}^t),
    \label{eq:2}
\end{equation}
\begin{equation}
    \Pr(\mathbf{P}^t | \mathbf{I}, \mathbf{P}^{t-1}, \hat{\mathbf{E}}^t) = \Pr(\widetilde{\mathbf{P}}^t | \mathbf{I}, \mathbf{P}^{t-1}, \hat{\mathbf{E}}^t) \cdot \Pr(\mathbf{P}^t | \hat{\mathbf{E}}^t, \widetilde{\mathbf{P}}^t), 
    \label{eq:3}
\end{equation}
\noindent where \(\mathbf{E}^t\) and \(\mathbf{P}^t\) represent the entity and predicate estimates at phase \(t\). \(\widetilde{}\) and \(\hat{}\) denote the intermediate estimates. The first terms of the two equations predict the intermediate estimates \(\widetilde{\mathbf{P}}^t\) and \(\hat{\mathbf{E}}^t\) based on the previous estimates and the image, corresponding to the first term of Eq. \ref{eq:3-1}. The second terms establish bidirectional dependencies between entities and predicates, corresponding to the second term of Eq. \ref{eq:3-1}. At each phase, these estimates will be updated through the corresponding decoder layer. The implementations of Eq. \ref{eq:2} and Eq. \ref{eq:3} are introduced as follows.

Inspired by previous work~\cite{li2022sgtr}, we initialize \(\mathbf{Q}_p^0\) randomly to generate the entity-aware predicate queries \(\mathbf{Q}_{com}^{t-1}\) for the current scene, implemented as follows:
\begin{equation}
    \mathbf{Q}_{com}^{t-1} = \mathcal{A}(q = \mathbf{Q}_p^0, k = \mathbf{V}_e, v = \mathbf{V}_e),
    \label{eq:}
\end{equation}
\noindent where \(\mathcal{A}\) denotes the attention block, and \(q, k, v\) represent the query, key, and value of attention networks, respectively. \(\mathbf{Q}_{com}^{t-1}\) serves as the input to the predicate branch at phase \(t\), consisting of three sub-queries: \(\mathbf{Q}_{io}^{t-1}\), \(\mathbf{Q}_p^{t-1}\), and \(\mathbf{Q}_{is}^{t-1}\). \(\mathbf{Q}_e^{t-1}\) is the input to the entity branch at phase \(t\), initialized from \(\mathbf{V}_e\). At phase \(t\), \(\mathbf{Q}_p^{t-1}\) and \(\mathbf{Q}_e^{t-1}\) are updated with \(\mathbf{V}\) via a cross-attention module, defined as follows:
\begin{equation}
    \hat{\mathbf{Q}}_e^t = \mathcal{A}(q = \mathbf{Q}_e^{t-1}, k = \mathbf{V}, v = \mathbf{V}),
    \label{eq:5}
\end{equation}
\begin{equation}
    \hat{\mathbf{Q}}_p^t = \mathcal{A}(q = \mathbf{Q}_p^{t-1}, k = \mathbf{V}, v = \mathbf{V}).
    \label{eq:6}
\end{equation}

The Eq. \ref{eq:5} corresponds to the first term of Eq. \ref{eq:2}. Then, \(\mathbf{Q}_{io}^{t-1}\) and \(\mathbf{Q}_{is}^{t-1}\) are updated with entity features \(\hat{\mathbf{Q}}_e^t\) through cross-attention. This step aims to identify entity pairs that match the corresponding predicates from the current entity detection. The process is implemented as follows:

\begin{equation}
    \mathbf{Q}_{io}^t = \mathcal{A}(q = \mathbf{Q}_{io}^{t-1}, k = \hat{\mathbf{V}}_e^t, v = \hat{\mathbf{V}}_e^t),
    \label{eq:7}
\end{equation}
\begin{equation}
    \mathbf{Q}_{is}^t = \mathcal{A}(q = \mathbf{Q}_{is}^{t-1}, k = \hat{\mathbf{V}}_e^t, v = \hat{\mathbf{V}}_e^t),
    \label{eq:7-1}
\end{equation}

\noindent where \(\hat{\mathbf{V}}_e^t = \mathbf{V}_e + \lambda \, \text{Norm}(\hat{\mathbf{Q}}_e^t)\). After updating these queries with the corresponding decoder, \(\hat{\mathbf{Q}}_p^t\) is further augmented with the updated indicator queries to adjust the predicate distribution, computed as follows:

\begin{equation}
    \widetilde{\mathbf{Q}}_p^t = (\hat{\mathbf{Q}}_p^t + (\mathbf{Q}_{io}^t + \mathbf{Q}_{is}^t) \cdot \mathbf{W}_i) \cdot \mathbf{W}_p.
    \label{eq:8}
\end{equation}

Eq. \ref{eq:6} to Eq. \ref{eq:8} correspond to the first term of Eq. \ref{eq:3}, where \(\mathbf{W}_i\) and \(\mathbf{W}_p\) represent transformation matrices. The indicator queries \(\mathbf{Q}_{io}^t\) and \(\mathbf{Q}_{is}^t\) are computed from entity features, enhancing the predicate query \(\widetilde{\mathbf{Q}}_p^t\) based on current entity detections. However, entity detection does not yet leverage information from predicates. To address this, we introduce a bidirectional attention module to establish conditional dependencies between entities and predicates, enabling mutual augmentation of entity and predicate features. This module is implemented as follows:

\begin{equation}
    \mathbf{Q}_e^t = \mathcal{A}(q = \hat{\mathbf{Q}}_e^t, k = \widetilde{\mathbf{Q}}_p^t, v = \widetilde{\mathbf{Q}}_p^t),
    \label{eq:9}
\end{equation}
\begin{equation}
    \mathbf{Q}_p^t = \mathcal{A}(q = \widetilde{\mathbf{Q}}_p^t, k = \hat{\mathbf{Q}}_e^t, v = \hat{\mathbf{Q}}_e^t).
    \label{eq:10}
\end{equation}

Eq. \ref{eq:9} and Eq. \ref{eq:10} correspond to the second terms of Eq. \ref{eq:2} and Eq. \ref{eq:3}, respectively. After the bidirectional interaction, the updated queries \(\mathbf{Q}_p^t\), \(\mathbf{Q}_{io}^t\), \(\mathbf{Q}_{is}^t\), and \(\mathbf{Q}_e^t\) are used as inputs for the next phase. Through multi-stage iterative refinement, the bidirectional interaction between entities and predicates is progressively enhanced. At the end of the iterations, the final queries \(\mathbf{Q}_{com}^{\text{end}}\) and \(\mathbf{Q}_e^{\text{end}}\) are fed into the corresponding Multi-Layer Perceptrons (MLPs) to predict the entity and predicate distributions, respectively.

\subsection{Random Feature Alignment}

In this subsection, we introduce the details of Random Feature Alignment, illustrated in Fig.~\ref{fig:3}. RFA distills knowledge from a pre-train CLIP model, constraining the feature space of various decoders aligned with CLIP. Furthermore, the parameters of the predicate and entity classifiers are initialized with the CLIP text encoder and fine-tuned on the SGG dataset. The specifics of feature distillation and classifier initialization are detailed as follows.

\subsubsection{Random Feature Alignment}
As illustrated in Fig.~\ref{fig:1} (a), the class imbalance in SGG datasets makes it challenging for BCG to capture feature interaction patterns for tail relationships from rare samples during training. Additionally, the learned feature interactions perform poorly on unseen relationships during inference. To address these issues, we propose RFA, which aligns the SGG feature space with that of a pre-trained vision-language model. In this semantically aligned feature space, BCG can effectively learn feature interaction patterns during training, and the learned patterns generalize better to unseen relationships.

\begin{figure*}[tb]
  \centering
  \includegraphics[width=1.0\linewidth]{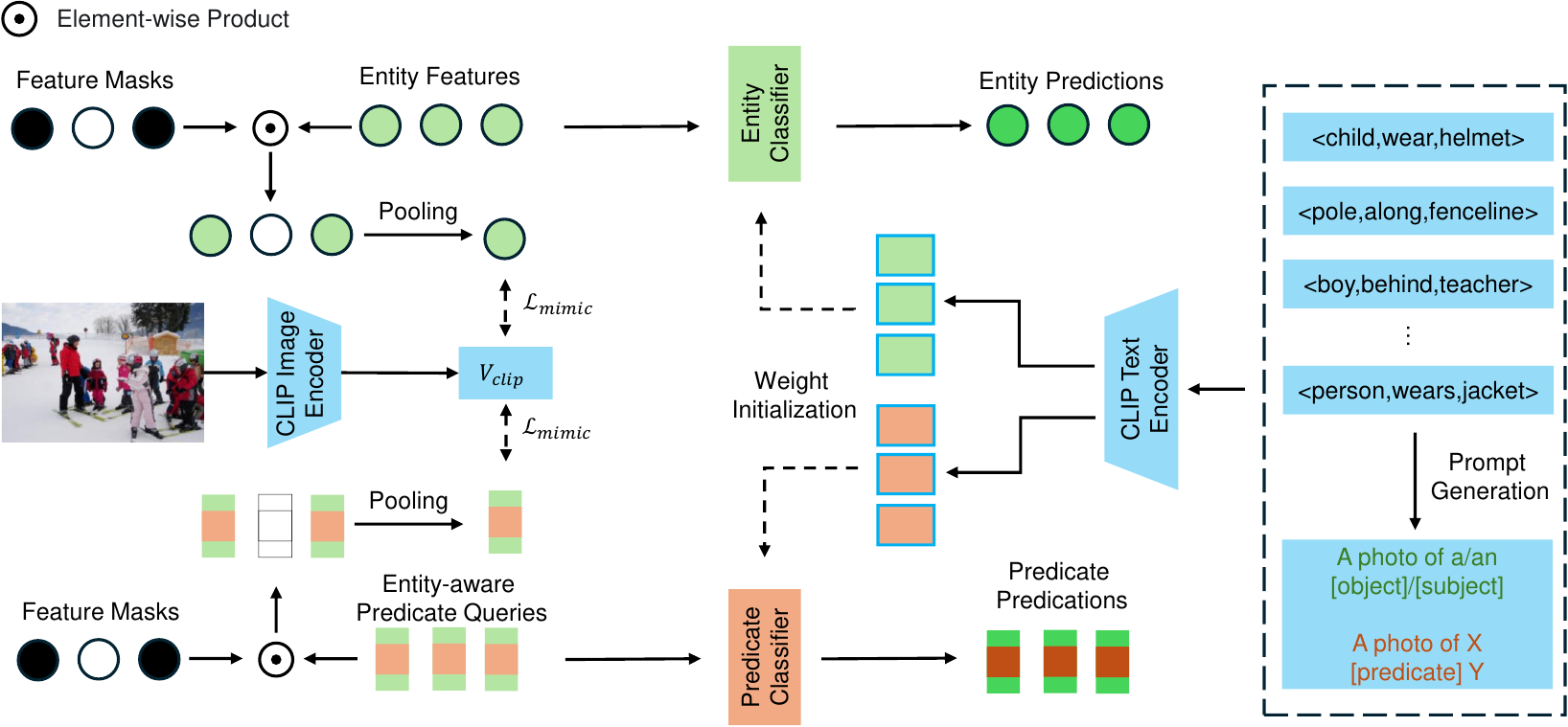}
  \caption{Random Feature Alignment for the entity and predicate prediction. First, the decoder features are randomly distilled with CLIP features. Then, the classifier weights are initialized with vectors generated by the CLIP text decoder, which encodes the ground-truth labels. This alignment ensures that the visual features can be accurately classified.
  }
  \label{fig:3}
\end{figure*}

Specifically, the input image is fed into the CLIP visual encoder to obtain \(\mathbf{V}_{clip}\). Since CLIP is trained with image captions, the encoder features \(\mathbf{V}_{clip}\) are inclined to capture the gist of the image. However, images often contain multiple relationships. Simply aligning the decoder's features \(\mathbf{Q}_p^{\text{end}}\) and \(\mathbf{Q}_e^{\text{end}}\) with the CLIP feature \(\mathbf{V}_{clip}\) may compromise the feature diversity of SGG. To preserve this diversity, we randomly select a subset of the decoder’s features before each alignment, ensuring that the distilled features retain diversity while being aligned with CLIP. The overall masked distillation process is as follows:

\begin{equation}
  \mathcal{L}_{mimic} = \left|\mathbf{V}_{clip} - \frac{1}{N_p}\sum_{i=1}^{N_p} \bar{\mathbf{Q}}_p^{\text{end}}\right| + \left|\mathbf{V}_{clip} - \frac{1}{N_e}\sum_{i=1}^{N_e} \bar{\mathbf{Q}}_e^{\text{end}}\right|,
  \label{eq:11}
\end{equation}

\noindent where \(\bar{\mathbf{Q}}_p^{\text{end}}\) and \(\bar{\mathbf{Q}}_e^{\text{end}}\) are sampled from \(\mathbf{Q}_p^{\text{end}}\) and \(\mathbf{Q}_e^{\text{end}}\) via randomly pooling. \(N_p\) and \(N_e\) denote the numbers of predicate and entity queries. An example is shown in Fig. \ref{fig:3}. Taking \(\mathbf{Q}_p^{\text{end}} \in \mathbb{R}^{N_e \times c}\) as an example, the mask vectors \(\mathbf{V}_{mask} \in \mathbb{R}^{N_e \times 1}\) are generated with a mask ratio \(\alpha\). The values of \(\mathbf{V}_{mask}\) are set to 0 with probability \(\alpha\), and the remaining values are set to \((1 / (1 - \alpha))\). Then, the dimensions of \(\mathbf{V}_{mask}\) are expanded and multiplied with \(\mathbf{Q}_p^{\text{end}}\) to produce \(\bar{\mathbf{Q}}_e^{\text{end}}\). Finally, the mean feature of \(\bar{\mathbf{Q}}_e^{\text{end}}\) is used to match the pre-trained features, as described in Eq. \ref{eq:11}.

Through random sampling features, some features are aligned with CLIP features, while others capture content that may be missing from CLIP features. After training, the feature space of the decoders aligns with CLIP’s feature space, enabling multiple queries to capture the rich triplet relationships within the image.

\subsubsection{CLIP-based Classifier}\label{sec_classifier}
To effectively utilize the aligned features \(\mathbf{Q}_p^{\text{end}}\) and \(\mathbf{Q}_e^{\text{end}}\), we introduce classifiers based on CLIP text features. Specifically, we generate descriptions by replacing \(*\) in templates like "A photo of a/an *" or "A photo of X * Y" with all the entity and predicate classes from the dataset. These descriptions are input to the CLIP text encoder to obtain the feature matrices \(\mathbf{W}_{\text{ent}} \in \mathbb{R}^{d \times D_e}\) and \(\mathbf{W}_{\text{pre}} \in \mathbb{R}^{d \times D_p}\), where $d$, $D_e$, and $D_p$ denote the dimension of CLIP features, the number of entity classes, and the number of predicate classes in the dataset, respectively. These matrices are used to initialize the last layer of the entity and predicate classifiers. Through the feature mapping and classifier layers, the decoder features \(\mathbf{Q}_p^{\text{end}}\) and \(\mathbf{Q}_e^{\text{end}}\) are projected and then compared with each category embedding via cosine similarity, from which the classification probabilities are computed. During training, both classifiers are fine-tuned with a small learning rate to improve performance on the SGG datasets. The entire process is illustrated in Fig. \ref{fig:3}.

\subsection{Graph Assembling}

This section outlines the process of generating final triplets using entity nodes \(\mathcal{V}_e\) and predicate nodes \(\mathcal{V}_p\). The process begins with the construction of adjacency matrices between \(\mathcal{V}_e\) and \(\mathcal{V}_p\), where the matrix elements are determined based on indicators of predicate and entity nodes. Specifically, two adjacency matrices \(\mathbf{M}_s\) and \(\mathbf{M}_o\) are computed. \(\mathbf{M}_o\) represents the relationships between subject nodes and predicate nodes, and \(\mathbf{M}_s\) represents the relationships between object nodes and predicate nodes. The relationship distances are derived using the corresponding predicted classes and bounding boxes. Taking the pair of objects and predicates as an example, given the object node set \(\mathcal{V}_o\) and the predicate node set \(\mathcal{V}_p\), we first construct the adjacency matrix \(\mathbf{M}_o\), which represents the distance between these nodes. The calculation is as follows:

\begin{equation}
  \mathbf{M}_o = d_{loc}(\mathbf{B}_o, \mathbf{B}_{po}) \cdot d_{cls}(\mathbf{P}_o, \mathbf{P}_{po}),
  \label{eq:dist}
\end{equation}

\noindent where \(d_{loc}\) and \(d_{cls}\) are distance functions used to measure matching quality of bounding box locations and classes. Specifically, \(\mathbf{B}_o\), \(\mathbf{B}_{po}\), \(\mathbf{P}_o\), and \(\mathbf{P}_{po}\) represent the bounding box predictions and classification distributions for objects and the indicators, respectively. Similarly, the adjacency matrix \(\mathbf{M}_s\) between subjects and predicates is calculated in the same way. For each predicate node in \(\mathcal{V}_p\), we select the top \(k\) edges from the two adjacency matrices \(\mathbf{M}_s\) and \(\mathbf{M}_o\). Consequently, a total of \(N_p \times k\) candidate triplets are generated as predictions in the form of \(T = \{(\mathbf{B}_o, \mathbf{B}_s, \mathbf{B}_p, \mathbf{P}_o, \mathbf{P}_s, \mathbf{P}_p)\}\), where \(\mathbf{B}_o, \mathbf{B}_s, \mathbf{B}_p\) and \(\mathbf{P}_o, \mathbf{P}_s, \mathbf{P}_p\) represent the bounding boxes and class predictions of its object, subject, and predicate, respectively.

\subsection{Training and Inference}
\subsubsection{Training}
To optimize the parameters of the proposed model, we design a multi-task loss function consisting of three components: \(\mathcal{L}_{ent}\) for the entity prediction, \(\mathcal{L}_{pre}\) for predicate prediction, and \(\mathcal{L}_{mimic}\) for feature distillation. The overall loss function is defined as follows:

\begin{equation}
  \mathcal{L} = \mathcal{L}_{ent} + \mathcal{L}_{pre} + \mathcal{L}_{mimic}.
  \label{eq:overloss}
\end{equation}

Since the entity detector follows a DETR-like architecture, \(\mathcal{L}_{ent}\) takes a similar form as described in~\cite{carion2020end}. To calculate \(\mathcal{L}_{pre}\), we first convert the ground-truth relationships in the image into the same format as predictions \(T\), denoted as \(T_{gt}\). Then, the Hungarian matching algorithm measures the cost between the ground truth and predictions, incorporating both predicate and entity information. The cost is computed as follows:

\begin{equation}
  C = \lambda_pC_p + \lambda_eC_e,
  \label{eq:hmcost}
\end{equation}

where $C_p$ and $C_e$ represent the costs for the predicate and entity, respectively. The matching results are obtained by selecting the minimum costs and are used to guide the calculation of $\mathcal{L}_{pre}$. Specifically, $\mathcal{L}_{pre}$ consists of two parts: $\mathcal{L}_{pre}^i$ and $\mathcal{L}_{pre}^p$, representing the losses for indicator and predicate, respectively. Both $\mathcal{L}_{pre}^i$ and $\mathcal{L}_{pre}^p$ include losses for bounding box predictions $\mathcal{L}_{box}$ (GIOU loss) and classification distributions $\mathcal{L}_{cls}$ (cross-entropy loss). The feature distillation loss $\mathcal{L}_{mimic}$ is implemented using the L1 loss.

\subsubsection{Inference} During inference, the feature distillation module is removed, as it is used only during training. After constructing the graph with predictions, we obtain $k \cdot N_p$ triplets. To produce the final results, we first filter out self-connected predictions where the object and subject in a triplet are identical. Next, we re-rank all triplets based on a belief score $S$ and select the top $k$ as final predictions. The belief score $S$ is calculated as the product of the classification probabilities of the corresponding subject entity, object entity, and predicate.

\section{Experiments}

\subsection{Datasets and Evaluation Metrics} 

\subsubsection{Visual Genome} Visual Genome~\cite{krishna2017visual} is the most representative SGG dataset, consisting of 57k training images, 5k validation images, and 26k test images. The dataset includes 150 object categories and 50 predicate categories. Following previous one-stage SGG studies~\cite{shit2022relationformer, im2024egtr}, we adopt the Scene Graph Detection (SGDET) setting to evaluate the proposed model. Following Tang et al.~\cite{tang2019learning}, we use mean Recall@k (mR@k) and Recall@k (R@k) as the evaluation metrics. mR@k represents the average recall across all classes, while R@k reflects the model's overall recall performance across categories. Inspired by previous work~\cite{zhang2022fine, min2023environment}, we use the overall metric F@K to jointly evaluate R@K and mR@K, which is the harmonic average of R@K and mR@K. Additionally, we report mR@100 for each category group (head, body, and tail) to assess the model's performance on long-tail distributions.

\subsubsection{Open Images V6} We also conduct experiments on the larger-scale Open Image V6~\cite{kuznetsova2020open} dataset, which consists of 126k training images, 5.3k test images, and 1.8k validation images. The dataset includes 288 object categories and 30 predicate categories. In addition to R@k, mR@k and F@k, we also report weighted mean average precision (AP) of phrase detection $wmAP_{phr}$ and relationship detection $wmAP_{rel}$. The $score_{wtd}$ is computed from R@k, $wmAP_{phr}$ and $wmAP_{rel}$ for a more class-balanced evaluation~\cite{cong2023reltr}.

\subsection{Implementation Details} We use ResNet-101 and DETR as the image feature extraction modules. The entity decoder adopts the same architecture as the DETR decoder. For Visual Genome \cite{krishna2017visual}, the numbers of entity nodes and predicate nodes are 100 and 160, respectively. For OpenImage V6 \cite{kuznetsova2020open}, these numbers are 100 and 180. The number of iterations (stages) is empirically set to 6. 
The value of $k$ is set to 25 for Visual Genome and 40 for OpenImage V6.

To ensure stable training and faster convergence, we first train the model using RFA.  The BCG is then incorporated into the training process for joint optimization. To train the model with RFA, we employ the AdamW optimizer with a weight decay rate of $1 \times 10^{-4}$. The entire model includes two fully connected layers followed by entity and predicate decoders, which map features into object and predicate category dimensions. The training process begins by optimizing the entity decoder on the target dataset using the entity feature distillation loss. Subsequently, the model is trained jointly with the feature distillation loss. The parameters of the entity decoder are frozen, while the predicate decoder remains trainable with a learning rate of $1 \times 10^{-4}$. Additionally, classifier parameters initialized from CLIP text features are updated with a small learning rate of $1 \times 10^{-5}$. The training process uses a batch size of 24 and runs for 90 epochs on 4 NVIDIA V100 GPUs. The loss function incorporates coefficients of 20, 1.0, and 0.5 for the RFA loss, entity detection loss, and predicate detection loss, respectively. 

During joint training, the BCG is integrated into the model. Specifically, an additional entity branch and a BiAtt module are introduced to enhance feature interaction. The entity decoder in BCG shares parameters with the entity decoder used during feature extraction, and these shared parameters are updated jointly during optimization. To ensure that BCG's feature interaction occurs in a semantically aligned space, we also compute the entity distillation loss at the entity decoder, enforcing constraints on the feature space during joint learning. The optimizer, learning rate, and batch size remain the same as in the first stage. The model is trained for 90 epochs to achieve final optimization.

\begin{table*}[t]
  \caption{\textbf{The Results Comparison in Visual Genome.}}
  \label{tab:1}
  \centering
  \scalebox{0.9}{
  \begin{tabular}{c|c|ccc|ccc|ccc|ccc}
    \toprule
    \multicolumn{2}{c|}{\multirow{2}{*}{\textbf{Method}}} &\multicolumn{3}{c|}{\textbf{mR@}} & \multicolumn{3}{c|}{\textbf{R@}} & \multicolumn{3}{c}{\textbf{mR@100}} & \multicolumn{3}{c}{\textbf{F@}}\\
    \multicolumn{2}{c|}{}&\textbf{20}&\textbf{50}&\textbf{100} & \textbf{20}&\textbf{50}&\textbf{100} & \textbf{Head}&\textbf{Body}&\textbf{Tail}& \textbf{20}&\textbf{50}&\textbf{100} \\
    \midrule
    \multirow{5}{*}{Two-stage} 
    & MOTIFS~\cite{zellers2018neural} & 4.2&5.7&- & 21.4&27.2&- & -&-&- &7.0&9.4&-\\
    & RelDN~\cite{zhang2019graphical} & -&6.0&7.3 & -&31.4&35.9 & -&-&- &-&10.1&12.1\\
    & VCTree-TDE~\cite{tang2020unbiased} & -&9.3&11.1 & -&19.4&23.2 & -&-&- &-&12.6&15.0\\
    & BGNN~\cite{li2021bipartite} & 7.5&10.7&12.6 & 23.3&31.0&35.8 & \textbf{34.0}&12.9&6.0 &11.3&15.9&18.6\\
    & RepSGG~\cite{liu2024repsgg} & 6.7&9.3&11.4 & 22.5&29.6&34.8 &31.3&11.2&5.3 &10.3&14.2&17.2\\
    \hline
    \multirow{11}{*}{One-stage} 
    & FCSGG~\cite{liu2021fully} & 2.7&3.6&4.2 & 16.1&21.3&25.1 & -&-&- &4.6&6.2&7.2\\
    & SRCNN~\cite{teng2022structured} & 6.2&8.6&10.3 & \textbf{26.1}&\textbf{33.5}&\textbf{38.4} & -&-&- &10.0&13.7&16.2\\
    & ISGG~\cite{khandelwal2022iterative} & -&8.0&8.8 & -&29.7&32.1 & 31.7&9.0&1.4 &-&12.6&13.8\\
    & Relationformer~\cite{shit2022relationformer} & 4.6&9.3&10.7 & 22.2&28.4&31.3 & -&-&- &7.6&14.0&15.9\\
    & SGTR~\cite{li2022sgtr} & -&12.0&15.2 & -&24.6&28.4 & 28.2&18.6&7.1 &-&16.1&19.8\\
    & RelTR~\cite{cong2023reltr} & 6.8&10.8&12.6 & 21.2&27.5&- & 30.6&14.4&5.0 &10.3&15.5&-\\
    & EGTR~\cite{im2024egtr} & 5.5&7.9&10.1 & 23.5&30.2&34.3 & -&-&- &8.9&12.5&15.6\\
    & PGSG~\cite{li2024pixels} & -&10.5&12.7 & -&20.3&23.6 & -&-&- &-&13.8&16.5\\
    & \textbf{BCTR (ours)} & \textbf{8.1}&\textbf{13.7}&\textbf{18.4} & 20.1&24.8&27.7 & 27.7&\textbf{22.0}&\textbf{11.9} &\textbf{11.5}&\textbf{17.6}&\textbf{22.1}\\
    \hline
    One-stage with & ISGG+Rw~\cite{khandelwal2022iterative} & -&15.7&17.8 & -&\textbf{27.2}&\textbf{29.8} & \textbf{28.5}&18.8&13.3 &-&\textbf{19.9}&22.3\\
    statistical-based & SGTR+Bilvl~\cite{li2022sgtr} & -&15.8&20.1 & -&20.6&25.0 & 21.7&21.6&17.1 &1.0&17.9&22.3\\
    long-tail  & RelTR+LA~\cite{cong2023reltr} & 9.7&14.2&- & \textbf{19.8}&25.9&- & 28.3&19.4&10.2 &13.0&18.3&-\\
     strategy & \textbf{BCTR (ours)+LA} & \textbf{12.7}&\textbf{17.4}&\textbf{20.9} & 17.2&21.9&25.2 & 24.6&\textbf{23.4}&\textbf{17.4}  &\textbf{14.6}&19.4&\textbf{22.8}\\
  \bottomrule
  \end{tabular}
  }
\end{table*}

\begin{table*}[t]
  \caption{\textbf{The Results Comparison in Open Image V6.} $\dag$ denotes our re-implementation following the setting in the paper.}
  \label{tab:2}
  \centering
  \scalebox{0.9}{
  \begin{tabular}{l|c|c|c|c|c|c}
    \toprule
    {\textbf{Method}}  & \textbf{R@50} & \textbf{$wmAP_{rel}$} & \textbf{$wmAP_{phr}$} & \textbf{$score_{wtd}$} & \textbf{mR@50} & \textbf{F@50}\\
    \hline
    RelDN~\cite{zhang2019graphical}&72.8&29.9&30.4&38.7&36.8&48.9\\
    GPS-Net~\cite{lin2020gps}&74.7&32.8&33.9&41.6&38.9&51.2\\
    BGNN~\cite{li2021bipartite} &\textbf{74.9}&31.2&31.4&40.0&39.4&51.6\\
    HOTR~\cite{kim2021hotr}&52.7&19.4&21.5&26.9&40.1&45.5\\
    AS-Net~\cite{chen2021reformulating}&55.3&25.9&27.5&32.4&35.2&43.0 \\
    SGTR$^{\dag}$~\cite{li2022sgtr} & 71.0 & 35.5 & 38.8 & 42.4& 44.1&54.4  \\
    PGSG~\cite{li2024pixels}  & 62.0 & 19.7 & 27.8 & 28.7& 40.7&49.1 \\
    \textbf{BCTR (ours)}  & 68.6 & \textbf{36.0} & \textbf{39.0} & \textbf{43.7}& \textbf{48.8} &\textbf{57.0}\\
  \bottomrule
  \end{tabular}
  }
\end{table*}

\subsection{Comparisons with State-of-the-Art Methods}

\subsubsection{Baselines} Since BCTR is a one-stage SGG method, we primarily compare it with other current one-stage methods, which are mainly transformer-based~\cite{li2022sgtr, khandelwal2022iterative, cong2023reltr, shit2022relationformer, he2023toward}. Additionally, we compare our method with several representative two-stage methods~\cite{li2021bipartite, tang2020unbiased, zellers2018neural, zhang2019graphical}. Although existing one-stage methods do not specifically tackle the long-tail problem, previous studies have reported results using statistical-based unbiased training (\textit{e.g.}, bi-level sampling~\cite{li2021bipartite}) or inference (\textit{e.g.}, logit adjustment~\cite{menon2020long}) techniques. For further comparison, we also report our method's performance with these statistical-based unbiased approaches.

\subsubsection{Results of Visual Genome} 
The experimental results in Tab.~\ref{tab:1} demonstrate that the BCTR model outperforms other baselines in mR@k performance. By leveraging BCG and RFA, BCTR learns balanced feature representations for scene graph generation, achieving significant improvements of 8\%, 14\% and 21\% in mR@k over the second-best method, indicating enhanced recall across various categories. In particular, BCTR excels in the body and tail categories, surpassing the second-best method by 18\% and 68\%, respectively. This performance is attributed to RFA, which constrains the feature space, enabling BCG to capture generalizable feature interactions even from rare samples, thereby boosting performance in tail categories.

BCTR's R@k appears lower because R@k and mR@k emphasize different aspects of model capability. In long-tailed datasets, the performance trade-off between head classes (categories with abundant data) and tail classes (categories with limited data) remains a central challenge~\cite{kangdecoupling}. Specifically, in the imbalanced VG dataset, R@k primarily reflects performance on head predicates, whereas mR@k highlights effectiveness on tail predicates. To this end, prior works~\cite{zhang2022fine, min2023environment} have introduced the F@k metric to evaluate R@k and mR@k jointly. Results demonstrate that BCTR achieves superior F@k scores compared to other methods, indicating its balanced and robust performance in the SGG task. Additionally, our DETR-based detector struggles with small entities which are common in the VG dataset~\cite{li2022sgtr}, leading to lower R@k. Improving small-object detection would enhance R@k. The experimental results in Tab.~\ref{tab:1} further confirm that, when combined with statistical long-tail strategies, BCTR surpasses other methods in the body and tail categories and significantly improves mR@k.

\subsubsection{Results of Open Image V6} The results on the Open Image V6 dataset are presented in Tab.~\ref{tab:2}. The experimental results show that BCTR achieves the highest performance across all class-balanced metrics and also gets comparable performance in R@50. In particular, BCTR outperforms the second-best method by 11\% in mR@50. The F@k results indicate that BCTR achieves better comprehensive performance on the Open Image dataset than other baselines. These results indicate that random feature alignment facilitates the bidirectional conditioning generator to effectively learn generalizable feature interaction patterns from imbalanced datasets, thereby enhancing the model's performance across various categories.

\subsection{Ablation Studies}

\begin{table}[t]
  \caption{\textbf{Ablation Study on Model Components. *Parameters are computed with 6-layer entity, predicate, and indicator decoders. FPS indicates the inference speed measured on a single V100 GPU. Epoch time refers to the duration of one training epoch on 8×V100 GPUs.}}
  \label{tab_exp_modelcom}
  \centering
  \scalebox{0.9}{
  \begin{tabular}{c|c|ccc|ccc|ccc|ccc}
    \toprule
    \multirow{2}{*}{\textbf{BCG}} & \multirow{2}{*}{\textbf{RFA}} &\multicolumn{3}{c|}{\textbf{mR@}} & \multicolumn{3}{c|}{\textbf{R@}} & \multicolumn{3}{c|}{\textbf{mR@100}} & \multirow{2}{*}{\textbf{Params*(M)}}& \multirow{2}{*}{\textbf{FPS}}& \multirow{2}{*}{\textbf{Epoch time (min)}}\\
    &&\textbf{20}&\textbf{50}&\textbf{100} & \textbf{20}&\textbf{50}&\textbf{100} & \textbf{Head}&\textbf{Body}&\textbf{Tail}& && \\
    
    \hline
    \XSolidBrush & \XSolidBrush & 7.4&12.8&17.0 & 19.6&24.2&27.3 & 26.9&21.3&9.7 & 96.1&5.1& 25.3 \\ 
    \Checkmark & \XSolidBrush & 7.7&13.2&17.6 & 19.9&24.6&27.6 & 27.2&21.5&10.9 & 99.6&3.8& 33.0 \\
    \XSolidBrush & \Checkmark & 7.9&13.2&17.9 & \textbf{20.4}&\textbf{25.1}&\textbf{28.2} & \textbf{28.1}&\textbf{22.4}&10.5 & 96.7&4.6&31.7 \\
    \Checkmark & \Checkmark & \textbf{8.1}&\textbf{13.7}&\textbf{18.4} & 20.1&24.8&27.7 & 27.7&22.0&\textbf{11.9} & 100.2&3.7&38.4 \\
  \bottomrule
  \end{tabular}
  }
\end{table}

\begin{table}[t]
  \caption{\textbf{Ablation Study on Distillation Strategy.}}
  \label{tab_exp_disstra}
  \centering
  \scalebox{0.9}{
  \begin{tabular}{c|c|c|ccc|ccc|ccc}
    \toprule
    \textbf{$\text{RFA}_\text{ENT}$} & \textbf{$\text{RFA}_\text{REL}$} & \textbf{Loss} & \textbf{mR@20} &\textbf{mR@50} & \textbf{mR100} & \textbf{R@20} & \textbf{R@50} & \textbf{R@100} & \textbf{Head}& \textbf{Body} & \textbf{Tail} \\
    \hline
    \XSolidBrush & \XSolidBrush & L1 & 7.3&12.9&17.6&	20.1&24.8&27.9&	27.6&22.0&10.2 \\
    \Checkmark & \XSolidBrush & L1&7.5&12.9&17.7&	20.4&25.3&28.5&	28.4&22.1&10.0 \\
    \XSolidBrush & \Checkmark & L1& 7.9&13.2&17.7&	20.4&25.3&28.4&	\textbf{28.6}&22.2&10.0 \\
    \hline
    \Checkmark & \Checkmark & L2&7.8&13.3&18.1&	20.4&25.2&28.3&	28.3&22.5&10.2 \\
    \Checkmark & \Checkmark & L1 & \textbf{7.9}&\textbf{13.6}&\textbf{18.3}&	\textbf{20.5}&\textbf{25.4}&\textbf{28.6}&	28.4&\textbf{22.6}&\textbf{11.0}\\
  \bottomrule
  \end{tabular}
  }
\end{table}

\subsubsection{Model Components} 
The results of the ablation study on model components are presented in Tab.~\ref{tab_exp_modelcom}, showing the performance of four model variants obtained by combining BCG and RFA. In Tab.~\ref{tab_exp_modelcom}, models with BCG include two additional decoder layers (i.e., bidirectional attention layers) compared to models without BCG. For a fair comparison, we add two extra decoder layers to the predicate decoder in the models without BCG.  The results demonstrate that both BCG and RFA contribute to improvements in mR@k and R@k.  Combining the two modules further improves recall in the tail category, resulting in a superior mR@100 performance of 18.4. We attribute this improvement to BCG's ability to learn better feature interactions within a semantically aligned space, which is regularized by RFA, thereby improving generalization to infrequent relationship combinations (i.e., tail categories). These results indicate that BCG and RFA are highly compatible, and integrating the two modules further enhances the performance of SGG.

The impacts of RFA and BCG on the model’s computational complexity and performance are reported in Tab.~\ref{tab_exp_modelcom}. Specifically, BCG and RFA introduce 3.5M and 0.6M additional parameters, respectively, resulting in an overall parameter increase of approximately 4.3\% compared to the baseline. During inference, BCG causes slower inference due to the added bidirectional feature interaction mechanisms, while RFA primarily functions during training and thus has negligible impact on inference speed. Regarding training performance, both components increase training time, with the overall training duration increases by 52\% when BCG and RFA are employed together. In summary, the increased computational cost of BCTR stems primarily from its architectural design rather than the number of additional parameters.

\subsubsection{Distillation Strategy}

The analysis of the effects of different distillation strategies on model performance is presented in Tab.~\ref{tab_exp_disstra}. We begin by evaluating the impact of distillation applied to different feature spaces. Specifically, we report the results of distilling only the entity features, only the relation features, and both simultaneously. The results indicate that the simultaneous distillation of both entity and relation features yields the best performance. Notably, distilling relation features alone achieves better performance than distilling entity features alone. We attribute this to the greater difficulty of relation detection compared to entity detection in the SGG task, whereby enhancing relation representations through distillation contributes more significantly to overall performance improvement.

In addition, we compare the effects of different mimic losses on distillation performance. As shown in Tab.~\ref{tab_exp_disstra}, the L1 loss yields better task performance than the L2 loss. This observation is consistent with the findings reported in \cite{liao2022gen}, which suggest that L1 loss is more suitable for feature distillation in relation detection tasks.

\subsubsection{Classifier} We conduct ablation experiments on distillation strategies to elucidate the effectiveness of the proposed RFA. The results are presented in Tab.~\ref{exp_tab_classifier}, where Trainable Classifier (TC) and Learnable Prompt (LP)~\cite{zhou2022learning} represent two CLIP-based classifiers. TC is the classifier introduced in Section \ref{sec_classifier}, which optimizes the parameters of the classifier during training. LP is the classifier introduced in~\cite{zhou2022learning}, which optimizes learnable prompts during training and performs classification through the text encoder of CLIP. The results indicate that Random Feature Alignment enhances the performance of both classifiers, particularly in terms of tail categories and mR@k. We attribute this improvement to RFA's ability to align decoder features with CLIP, enabling the features of tail categories to be well-represented and correctly classified, even if these categories are infrequent during training. Furthermore, the results demonstrate that TC outperforms LP in the scene graph generation task.

\begin{table}[t]
  \caption{\textbf{Ablation Study on Classifier.} Trainable Classifier (TC) and Learnable Prompt (LP)~\cite{zhou2022learning} represent two different types of classifiers based on CLIP features.}
  \label{exp_tab_classifier}
  \centering
  \scalebox{0.9}{
  \begin{tabular}{c|ccc|ccc|ccc}
    \toprule
    \textbf{Method} & \textbf{mR@20} &\textbf{mR@50} & \textbf{mR100} & \textbf{R@20} & \textbf{R@50} & \textbf{R@100} & \textbf{Head}& \textbf{Body} & \textbf{Tail} \\
    \hline
    TC & 7.8&13.0&17.8 & 20.1&25.1&\textbf{28.3} & \textbf{28.2}&22.4&10.0 \\
    TC-RFA & \textbf{7.8}&\textbf{13.3}&\textbf{18.2} & \textbf{20.4}&\textbf{25.1}&28.2 & 28.1&\textbf{22.5}&\textbf{10.9} \\
    \hline
    LP & 7.5&13.0&17.2 & 20.1&\textbf{25.3}&\textbf{28.7} & \textbf{28.4}&\textbf{22.2}&8.9 \\
    LP-RFA & \textbf{7.5}&\textbf{13.0}&\textbf{17.6} & \textbf{20.2}&25.2&28.3& 28.0&21.9&\textbf{10.2} \\
  \bottomrule
  \end{tabular}
  }
\end{table}

\begin{table}[t]
  \caption{\textbf{Experiments on Zero-shot Triplets Generation.}}
  \label{tab:6}
  \centering
  \scalebox{0.9}{
  \begin{tabular}{l |cc}
    \toprule
    \textbf{Method} & \textbf{zR@50} & \textbf{zR@100} \\
    \hline
    BGNN~\cite{li2021bipartite}&0.4&0.9\\
    VCTree-TDE~\cite{tang2020unbiased}&2.6&3.2\\
    SGTR~\cite{li2022sgtr}&2.4&5.8\\
    ISGG~\cite{khandelwal2022iterative}&3.9&5.6\\
    \hline
    \textbf{BCTR (ours) w/o BCG+RFA}&  3.6&5.1 \\
    \textbf{BCTR (ours) w/ BCG} & 3.7&5.4 \\
    \textbf{BCTR (ours) w/ RFA} & 4.1&6.0 \\
    \textbf{BCTR (ours) w/ BCG+RFA} & \textbf{4.4}&\textbf{6.2} \\
  \bottomrule
  \end{tabular}
  }
\end{table}

\subsubsection{Zero-shot Recall} The Zero-shot Recall@k (zR@k) of different models is reported in Tab.~\ref{tab:6}. zR@k measures the recall of triplets that are unseen during training (i.e., combinations of subject, predicate, and object that does not appear during training), reflecting the model's generalization performance. The results show that the Bidirectional Conditioning Generator provides a minor improvement in zR@k, with zR@50/100 increasing by 0.1 and 0.3, respectively. When combined with Random Feature Alignment, zR@k improves significantly to 4.4 and 6.2, exceeding the second-best method by 13\% and 7\%, respectively. This finding aligns with the observations in Tab.~\ref{tab_exp_modelcom}, suggesting that feature interactions learned in semantically aligned spaces generalize more effectively to unseen category relationships.

\begin{table}[t]
  \caption{\textbf{Ablation Study on Mask Ratio.}}
  \label{tab:7}
  \centering
  \scalebox{0.9}{
  \begin{tabular}{c|ccc|ccc|ccc}
  \toprule
    \textbf{Mask Ratio} & \textbf{mR@20} &\textbf{mR@50} & \textbf{mR100} & \textbf{R@20} & \textbf{R@50} & \textbf{R@100} & \textbf{Head}& \textbf{Body} & \textbf{Tail} \\
    \hline
    0 & 7.8&13.0&17.8 & 20.1&\textbf{25.1}&\textbf{28.3} & \textbf{28.2}&22.4&10.0 \\
    0.25 & 7.8&13.2&17.8 & 20.2&24.9&27.9 & 27.8&22.5&10.2 \\
    0.5 & \textbf{7.8}&\textbf{13.3}&\textbf{18.2} & \textbf{20.4}&25.1&28.2 & 28.1&\textbf{22.5}&\textbf{10.9} \\
    0.75 & 7.5&13.3&17.9 & 20.2&24.8&28.0&  28.0&22.4&10.5 \\
  \bottomrule
  \end{tabular}
  }
\end{table}

\subsubsection{Mask Ratio} We set the mask ratio $\alpha$ between 0 and 1 to evaluate its effect on model performance. A larger value of $\alpha$ indicates that the model has conducted more thorough feature distillation during training. In this case, the feature spaces of entities and predicates are aligned with CLIP features, but some detailed features that are not encoded by CLIP may be lost. Conversely, a smaller $\alpha$ means that the feature spaces of entities and predicates are more aligned with the dataset. The experimental results are presented in Tab.~\ref{tab:7}. The R@k is high when $\alpha$ is close to 0, as the model captures the main content of the image after aligning with the CLIP feature space, which accounts for a higher proportion of the VG dataset. As $\alpha$ increases, the model captures more diverse content, leading to improvements in mR@k. However, when $\alpha$ approaches 1, performance declines due to insufficient feature distillation. The experimental results show that the proposed model achieves the best performance when $\alpha$ is set to 0.5.

\begin{table}[t]
  \caption{\textbf{Experiments on different image resolutions.}}
  \label{tab:9}
  \centering
  \scalebox{0.9}{
  \begin{tabular}{c|ccc|ccc|ccc}
    \toprule
    \textbf{Reslution} & \textbf{mR@20} &\textbf{mR@50} & \textbf{mR100} & \textbf{R@20} & \textbf{R@50} & \textbf{R@100} & \textbf{Head}& \textbf{Body} & \textbf{Tail} \\
    \hline
    400$\times$400&7.3&12.7&16.6&18.1&22.3&24.9  & 25.2&20.6&9.9  \\
    500$\times$500&7.9&13.6&17.9&19.8&24.4&27.2 & 27.1&21.9&11.1  \\
    600$\times$600&\textbf{8.1}&\textbf{13.7}&\textbf{18.4}&\textbf{20.1}&\textbf{24.8}&\textbf{27.7} & \textbf{27.7}&\textbf{22.0}&\textbf{11.9}\\
    700$\times$700&8.0&13.5&18.0&20.0&24.6&27.4 & 27.4&21.4&11.6\\
    800$\times$800&8.0&13.4&17.9&19.9&24.5&27.3 & 27.2&21.7&11.3\\
  \bottomrule
  \end{tabular}
  }
\end{table}

\begin{table}[t]
  \caption{\textbf{Experiments on different scenarios.}}
  \label{tab:8}
  \centering
  \scalebox{0.9}{
  \begin{tabular}{l|c|cc|cc}
    \toprule
    \textbf{Model}&\textbf{Scene} & \textbf{mR@20}&\textbf{mR@50} & \textbf{R@20}& \textbf{R@50}\\
    \hline
        SGTR~\cite{li2022sgtr}&Simple & 7.8&12.4&19.2&23.9\\
     \textbf{BCTR (ours)}&Simple & \textbf{8.8}&\textbf{14.1} & \textbf{19.8}&\textbf{24.2}\\
    \hline
        SGTR~\cite{li2022sgtr}&Normal &7.4&12.8&19.9&25.0\\
     \textbf{BCTR (ours)}&Normal & \textbf{9.0}&\textbf{14.1}&\textbf{20.5}&\textbf{25.4}\\
    \hline
        SGTR~\cite{li2022sgtr}&Complex &7.3&12.2&19.5&24.7 \\
     \textbf{BCTR (ours)}&Complex & \textbf{7.5}&\textbf{13.3}&\textbf{19.9}&\textbf{25.0}\\

  \bottomrule
  \end{tabular}
  }
\end{table}

\subsubsection{Image Resolution} We conducted an assessment of the trained model's generalization across different image resolutions. Initially, the base model is trained on images with a resolution of 600$\times$600. Subsequently, the model's performance is evaluated on images of various resolutions. The experimental results are presented in Table~\ref{tab:9}. The results indicate that performance degrades as the image resolution deviates from the baseline resolution of 600$\times$600. The performance drop is more pronounced at lower resolutions, primarily because the DETR-based detector struggles to effectively detect small objects at reduced scales. To improve model robustness to varying input image resolutions in real-world deployment scenarios, techniques such as multi-scale random scaling during training~\cite{tian2023resformer}, multi-scale ensemble inference~\cite{najibi2019autofocus}, and auxiliary super-resolution modules~\cite{zhang2023superyolo} can be employed.

\subsubsection{Content Complexity} We compare the performance of BCTR across diverse scenarios. Specifically, we analyze the number of triplets within the images of the VG test dataset. Subsequently, we partitioned the test set into three equal-sized subsets based on the number of triplets in the images. The Simple scene set includes images with fewer than 4 triplets; the Normal scene set includes images with 4 to 8 triplets; and the Complex scene set includes images with more than 8 triplets. The performance of different models in various scenarios is presented in Table~\ref{tab:8}. The results demonstrate that our method consistently outperforms the baseline across varying levels of complexity, including highly complex scenarios.

\subsection{Qualitative Analysis} 
\begin{figure*}[!t]
  \centering
  \includegraphics[width=1.0\linewidth]{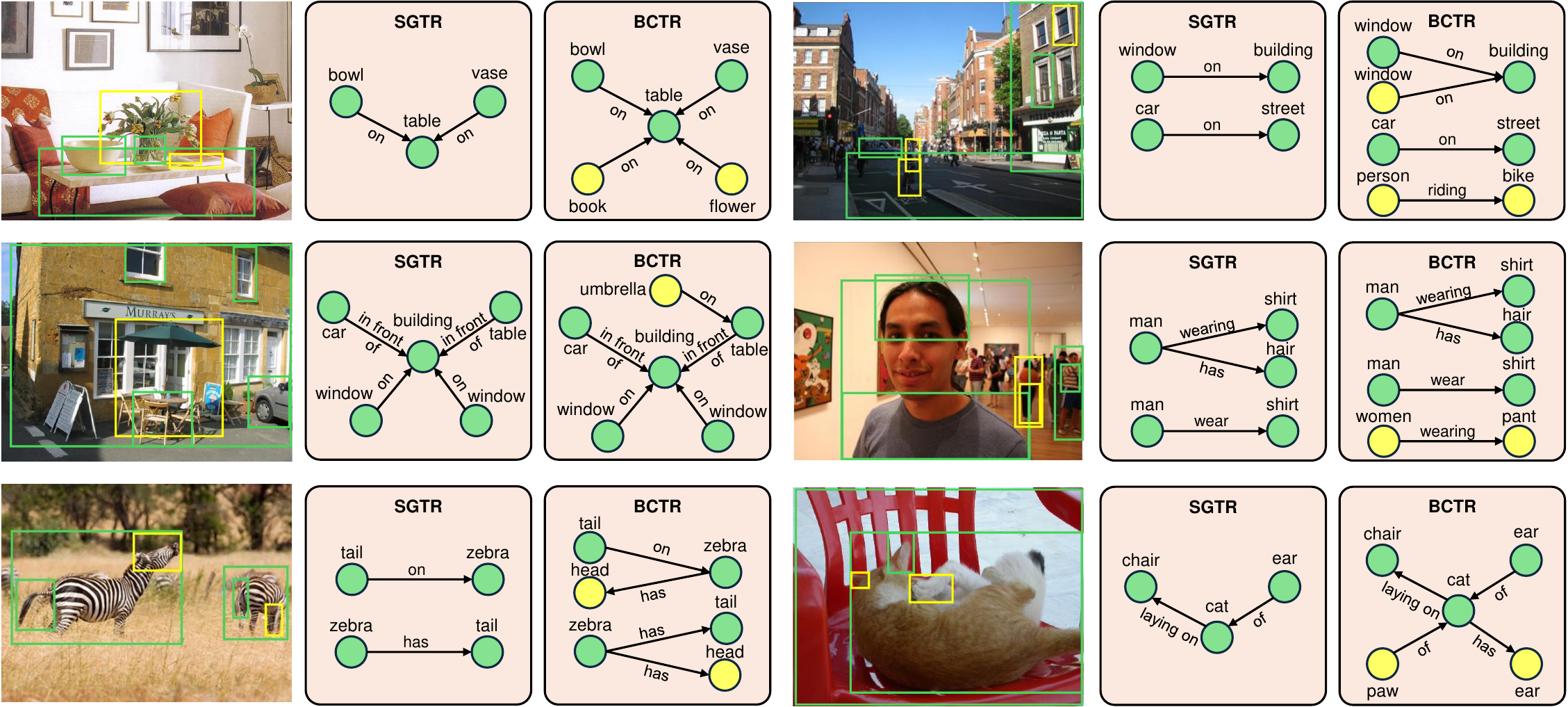}
  \caption{Qualitative results of our method and another method on the VG dataset. When leveraging identical DETR-based detectors, the bidirectional interaction mechanism of BCTR reduces missed detections (highlighted as yellow nodes) and enhances performance on the SGG task.
  }
  \label{fig:4}
\end{figure*}

We visualize the detection results of the VG dataset in Fig.~\ref{fig:4}. The test images cover various scenarios such as indoor, outdoor, human, animals, etc. For clarity, only the directed edges matching the ground-truth are shown. The experimental results show that BCTR detects more accurate relationships than the baseline, whether they are prominent relationships (e.g., umbrella on table, flower on table) or detailed relationships (e.g., women wearing pant, person riding bike) in the images. It is worth noting that SGTR~\cite{li2022sgtr} and BCTR adapt the same detector, yet BCTR achieves better results. This suggests that the bidirectional interaction mechanism of BCTR mutually enhances entity and predicate detection, leading to superior performance.

\section{Conclusion}
In this study, we propose a novel bidirectional conditioning factorization in semantic-aligned space for SGG and implement it by developing an end-to-end SGG model BCTR. BCTR enhances the performance of SGG tasks in two main aspects. First, the Bidirectional Conditioning Generator is designed to facilitate information interaction between entity and predicate predictions through internal bidirectional conditioning and external iterations. Second, Random Feature Alignment is introduced to facilitate BCG in learning interaction patterns by randomly aligning the feature space with a pre-trained visual language model, improving generalization for semantically related relationships during inference. We perform experiments on several datasets: Visual Genome and Open Images V6, and the results demonstrate that our method achieves state-of-the-art performance. 

The limitation of the proposed method lies in the ongoing challenges associated with distilling features from VLPMs to SGG. Firstly, while RFA is introduced to address the discrepancy between CLIP's focus on capturing global features and the SGG task's focus on local predicates, exploring more fine-grained VLPM enhancements could be beneficial for improving performance in the future. Additionally, CLIP exhibits poor performance in capturing structured knowledge, such as object properties and predicates, which are crucial components in SGG tasks. Therefore, exploring feature distillation with more advanced CLIP models is an interesting direction for further investigation.




\section*{Acknowledgments}
This work was supported in part by the National Key Research and Development Program of China under Grant 2023YFB4705000, in part by the National Natural Science Foundation of China under 62303455, 62373039, U23B2038, U24A20282, 62273342, 62403461, and 62122087, in part by Beijing Natural Science Foundation under Grant L233006, in part by the Youth Program and the Open Projects Program of State Key Laboratory of Multimodal Artificial Intelligence Systems.



\bibliographystyle{cas-model2-names}

\bibliography{cas-refs}

\end{document}